\documentclass[conference]{IEEEtran}
\IEEEoverridecommandlockouts
\usepackage{cite}
\usepackage{amsmath,amssymb,amsfonts}
\usepackage{algorithmic}
\usepackage{graphicx}
\usepackage{textcomp}
\usepackage{xcolor}

\usepackage{hyperref}
\usepackage{cite}
\usepackage{tcolorbox}
\usepackage{tabularx}
\usepackage{booktabs}
\usepackage{enumitem}

\def\BibTeX{{\rm B\kern-.05em{\sc i\kern-.025em b}\kern-.08em
    T\kern-.1667em\lower.7ex\hbox{E}\kern-.125emX}}
\begin{document}

\title{Optimizing Large Language Models: Metrics, Energy Efficiency, and Case Study Insights
}





\author{

\IEEEauthorblockN{Tahniat Khan}
\IEEEauthorblockA{\textit{Industry Innovation, Vector Institute} \\
Toronto, Canada \\
tahniat.khan@vectorinstitute.ai}
\and
\IEEEauthorblockN{Soroor Motie}
\IEEEauthorblockA{\textit{Vector Institute, University of Ottawa} \\
Ottawa, Canada \\
smoti088@uottawa.ca} 
\and
\IEEEauthorblockN{Sedef Akinli Kocak}
\IEEEauthorblockA{\textit{Industry Innovation, Vector Institute} \\
Toronto, Canada \\
sedef.kocak@vectorinstitute.ai}

\and

\IEEEauthorblockN{\centerline{Shaina Raza}}
\IEEEauthorblockA{\textit{AI Engineering, Vector Institute} \\
Toronto, Canada \\
shaina.raza@vectorinstitute.ai}
}

\IEEEoverridecommandlockouts
\IEEEpubid{\makebox[\columnwidth]{\copyright~2025 IEEE. Accepted to IEEE CAI 2025, to appear in IEEE Xplore. \hfill}
\hspace{\columnsep}\makebox[\columnwidth]{ }}

\maketitle

\begin{abstract}

The rapid adoption of large language models (LLMs) has led to significant energy consumption and carbon emissions, posing a critical challenge to the sustainability of generative AI technologies. This paper explores the integration of energy-efficient optimization techniques in the deployment of LLMs to address these environmental concerns. We present a case study on sentiment analysis and framework that demonstrate how strategic quantization and local inference techniques can substantially lower the carbon footprints of LLMs without compromising their operational effectiveness. Experimental results reveal that these methods can reduce energy consumption and carbon emissions by up to 55\% post quantization, making them particularly suitable for resource-constrained environments. The findings provide actionable insights for achieving sustainability in AI while maintaining high levels of accuracy and responsiveness.

\end{abstract}

\begin{IEEEkeywords}
Large Language Models (LLMs), Quantization, Green AI, Carbon Emissions, Energy Efficiency
\end{IEEEkeywords}

\section{Introduction}
The increasing computational demands of advanced artificial intelligence (AI), particularly generative models including large language models (LLMs), have motivated significant research and development in \emph{Green AI}. It highlights the importance of adopting sustainable practices to mitigate the rising environmental impact of generative AI technologies \cite{bolon2024review}. However, despite these advances, many generative AI applications continue to consume substantial computational resources, leading to increased energy consumption and elevated carbon emissions \cite{verdecchia2023systematic}. As these generative AI models and applications scale in both size and complexity, they demand frequent data and model updates, creating a potentially unending cycle of energy-intensive processes that could hinder overall progress in sustainable AI \cite{verdecchia2023systematic}.

Generative AI systems such as ChatGPT, GPT-3, Claude, and Llama demonstrate remarkable capabilities in text generation and reasoning but come with significant environmental costs associated with their development and infrastructure \cite{bolon2024review}. Data centers, which support the underlying computational needs of these tools, account for ~1–1.5\% of global electricity consumption and <1\% of energy‑related CO2e, \footnote{https://www.iea.org/reports/energy-and-ai/executive-summary} with rapid growth expected due to AI workloads and data volumes. Moreover, hyperscale cloud providers, including Amazon AWS, Google Cloud, and Microsoft Azure, commonly use power-intensive GPUs for hosting generative AI models; these GPUs can consume 10–15 times the energy of traditional CPUs, significantly enlarging the technology’s carbon footprint \cite{barbierato2024towards}. Understanding the full lifecycle of emissions for machine learning models is therefore essential for tackling these environmental challenges \cite{barbierato2024towards}. Strategies aimed at lowering energy demands across this lifecycle—ranging from pre-training to inference—represent a critical step in achieving sustainable generative AI solutions \cite{barbierato2024towards}.

\paragraph{Motivation}

Motivated by the growing environmental impact of large language models (LLMs), this study focuses on optimizing the inference stage, where energy use and carbon emissions can be significant during deployment. While there is increasing recognition of the need for energy-efficient LLMs, there is still limited awareness and few practical demonstrations showing how optimization can reduce environmental costs without sacrificing performance. Addressing this gap requires not only quantifying the carbon footprint of inference but also evaluating concrete optimization strategies. Furthermore, analyzing a specific use case provides valuable insights into how energy-efficient methods can be adopted in practice while preserving model effectiveness.

\paragraph{Objectives}

In this work, we examine the metrics and units currently used to measure the environmental footprint of LLMs during inference and evaluate how these metrics change when applying optimization techniques. We then present a case study in sentiment analysis that demonstrates how targeted optimizations can make inference more energy efficient without compromising accuracy.

The primary objectives of this study are threefold:
\begin {itemize} 
\item To enhance the energy efficiency of LLMs by measuring and minimizing energy consumption during inference.

\item To reduce the associated carbon emissions by applying and assessing optimization strategies for inference workloads.

\item To develop and evaluate a methodology that prioritizes performance preservation, ensuring accuracy and responsiveness remain high while achieving measurable sustainability gains.

\end {itemize} 

\paragraph{Contributions}

This study makes three main contributions to the growing field of Green AI. First, it presents an evaluation framework for quantifying the energy use and carbon footprint of LLMs during inference, a stage critical to real-world deployment. Second, it implements and assesses an optimization framework combining quantization and local inference, showing how these techniques reduce energy usage and emissions. Third, it provides a detailed case study on sentiment analysis, offering empirical evidence that optimization can reduce emissions by up to 45\%  with only minimal impact on model performance. Together, these contributions provide practical insights for advancing sustainability in AI deployment while maintaining robust performance.

\section{Related Work}

In recent years, the convergence of environmental sustainability and AI has led to the emergence of ``in \emph{Green AI}", focusing on reducing the carbon footprint of 
large-scale models through optimization techniques. This section reviews key studies that have contributed to this field, highlighting their contributions to sustainable AI practices.

Efforts to mitigate the environmental impact of LLMs have focused on understanding and reducing their carbon footprint. Studies have quantified the CO$_2$ emissions associated with large-scale models \cite{bolon2024review, barbierato2024towards}, highlighting significant environmental challenges posed by their extensive parameter sizes and computational demands \cite{liu2024green}. Liu and Yin (2024), in particular, emphasizes the critical role of hardware choices in sustainable AI practices and proposes training methods without compromising performance to reduce carbon emissions \cite{liu2024green}. These foundational insights underscore the urgency of addressing sustainability in LLM development and deployment.

Building on this foundation, several tools and frameworks have been proposed. For example, GreenTrainer \cite{huang2023towards} has been introduced as a fine-tuning approach that dynamically evaluates backpropagation costs and contributions to model accuracy. By reducing floating-point operations (FLOPs) during fine-tuning by up to 64\%, GreenTrainer achieves significant energy savings without compromising model performance \cite{huang2023towards}. Likewise, Avatar focuses on creating compact, energy-efficient models optimized for deployment on individual devices \cite{shi2024greening}. By reducing inference latency and model size, this method significantly decreases the carbon footprint of LLM usage while maintaining competitive performance \cite{raza2024developing}, illustrating the potential of targeted optimizations \cite{luccioni2023estimating}.

In addition to specific optimization techniques, broader frameworks for sustainable AI have been proposed. These frameworks advocate for the integration of energy-efficient algorithms and the alignment of AI practices with global sustainability goals \cite{tabbakh2024towards}. For instance, intersection of sustainability and software engineering is well established research area \cite{duboc2019we, venters2021software, betz2024lessons} such as exploring strategies for creating eco-friendly solutions that maintain functionality while reducing energy consumption \cite{shi2024efficient}. Collectively, these studies provide actionable insights and a roadmap for advancing Green AI, addressing immediate environmental concerns, and fostering a sustainable future for AI technologies.

\subsection{Carbon Emission Metrics}

Measuring carbon emissions is essential for understanding and reducing the environmental impact of AI systems. Various metrics have been developed to assess emissions across different scopes, intensities, and stages. This subsection outlines widely used metrics, such as Carbon Dioxide Equivalent (CO$_2$e), Carbon Intensity, and Global Warming Potential (GWP) and highlights their role in promoting transparency and sustainability (Table~\ref{tab:emission_metrics}).


\begin{table}[ht]
\caption{Common Carbon Emission Metrics in Green AI}
\label{tab:emission_metrics}
\centering
\footnotesize
\begin{tabular}{|p{2.2cm}|p{1cm}|p{2.8cm}|p{1.4cm}|}
\hline
\textbf{Metric} & \textbf{Unit} & \textbf{Definition} & \textbf{Reference} \\
\hline
Carbon Dioxide Equivalent (CO\textsubscript{2}e) & Metric tons (tCO\textsubscript{2}e) & A measure of greenhouse gases expressed as CO\textsubscript{2} equivalent & \href{https://www.ipcc.ch/}{IPCC}, \href{https://ghgprotocol.org/}{GHG Protocol} \\
\hline
Carbon Intensity & gCO\textsubscript{2}/ kWh & CO\textsubscript{2} emissions per unit of electricity consumed & \href{https://www.iea.org/}{International Energy Agency} \\
\hline
Scope 1 Emissions & tCO\textsubscript{2}e & Direct emissions from controlled sources & \href{https://ghgprotocol.org/}{GHG Protocol} \\
\hline
Scope 2 Emissions & tCO\textsubscript{2}e & Indirect emissions from purchased electricity & \href{https://ghgprotocol.org/}{GHG Protocol} \\
\hline
Scope 3 Emissions & tCO\textsubscript{2}e & Indirect emissions across value chains & \href{https://ghgprotocol.org/}{GHG Protocol} \\
\hline
Net Zero Emissions & tCO\textsubscript{2}e & Balance when emissions equal removals & \href{https://unfccc.int/}{UNFCCC} \\
\hline
Energy Consumption & MWh & Total energy consumed & \href{https://www.iea.org/}{IEA}, \href{https://www.eia.gov/}{EIA} \\
\hline
Global Warming Potential (GWP) & Ratio & Heat trapped by a gas compared to CO\textsubscript{2} & \href{https://www.ipcc.ch/}{IPCC} \\
\hline
Carbon Offsets & tCO\textsubscript{2}e & Credits for emissions reduction or removal & \href{https://verra.org/}{VCS}, \href{https://www.goldstandard.org/}{Gold Standard} \\
\hline
Carbon Capture and Storage (CCS) & tCO\textsubscript{2} captured & CO\textsubscript{2} removed and stored to prevent release & \href{https://www.iea.org/}{IEA}, \href{https://www.ipcc.ch/}{IPCC} \\
\hline
\end{tabular}
\end{table}

\subsection{Quantization Techniques in LLMs}

Quantization \cite{han2015deep} has emerged as a transformative approach in optimizing LLMs, addressing the dual challenges of computational efficiency and environmental sustainability. It works by converting model parameters from high-precision formats (e.g., 32-bit floating-point) to lower-precision formats (e.g., 8-bit or even 4-bit), thereby reducing memory requirements and accelerating computation. This technique aligns closely with the goals of Green AI, as it minimizes resource usage while maintaining acceptable accuracy. 

Research efforts have showcased the potential of quantization as a key technique for enhancing energy efficiency in AI systems. For instance, GPTQ (Accurate Post-Training Quantization for Generative Pre-trained Transformers) \cite{frantar2022gptq} introduces a method for post-training quantization that retains high model performance despite significant reductions in parameter precision. This method enables efficient deployment of LLMs on edge devices and other constrained environments, directly addressing the environmental concerns highlighted in studies like GreenTrainer \cite{huang2023towards}. Another notable approach is LLM-QAT (Quantization-Aware Training for Large Language Models) \cite{liu2023llm}, which integrates quantization during the training phase rather than applying it post-training. This method further improves the trade-off between model size and performance, making LLMs more adaptable to energy-efficient deployments. Additionally, SmoothQuant  \cite{liu2023llm} employs layer-wise quantization to balance computation and accuracy, achieving state-of-the-art results in reducing energy use during inference. 

\subsection{Balancing Efficiency Gains with Minimal Performance Impact}
The relationship between optimization and predictive performance in LLMs demonstrates that substantial efficiency gains can be achieved without meaningful sacrifices in accuracy. While optimization consistently reduces carbon emissions by 25–55\%, the corresponding impact on metrics such as precision, recall, F1 score, and overall accuracy is minimal, indicating that performance remains largely preserved even as resource efficiency improves. Techniques like FrugalGPT, described by Chen et al. (2023) \cite{chen2023frugalgpt}, illustrate how cascading models and leveraging prompt adaptation can reduce costs by up to 98\% without compromising accuracy. Similarly, FrugalML shows how selectively routing queries to different APIs can maintain performance while cutting costs by up to 90\% \cite{chen2023frugalgpt, chen2020frugalml}. However, the effectiveness of these strategies varies across tasks. For example, reducing token lengths or approximating model outputs might preserve general quality but risks performance drops in nuanced applications like sentiment analysis or summarization \cite{shekhar2024towards}. These findings reveal that while cost and carbon footprint reductions are achievable, ensuring minimal trade-offs in precision, recall, or F1 score remains a complex optimization problem, often requiring task-specific calibrations \cite{chen2023frugalgpt, shekhar2024towards}.

\section{Case Study: Sustainable Deployment of Large Language Models}

\subsection{Problem Definition}
LLMs have become integral to various natural language processing applications, yet their soaring computational demands pose significant sustainability challenges. These include high energy consumption, carbon emissions, and escalating operational costs, particularly when using cloud-based infrastructures \cite{bolon2024review}. To address these concerns, this study proposes a framework for LLM deployment that emphasizes \emph{local inference}, aiming to mitigate environmental impact while preserving model performance and user experience.

Formally, consider a classification problem with input data 

\begin{equation}
X = \{ x_1, x_2, \ldots, x_N \}\label{eq},
\end{equation}

and corresponding ground-truth labels 

\begin{equation}
Y = \{ y_1, y_2, \ldots, y_N \}, \quad y_i \in \{1, 2, \dots, K\}\label{eq},
\end{equation}

An LLM-based classifier \( f_\theta(\cdot) \) predicts \(\hat{y}_i = f_\theta(x_i)\). We seek to minimize energy consumption and carbon emissions while maintaining high predictive accuracy, where accuracy can be quantified using standard metrics such as precision, recall, and F1-score.

\subsection{Framework Overview}
\begin{figure}[htbp]
\centering
\includegraphics[width=\columnwidth]{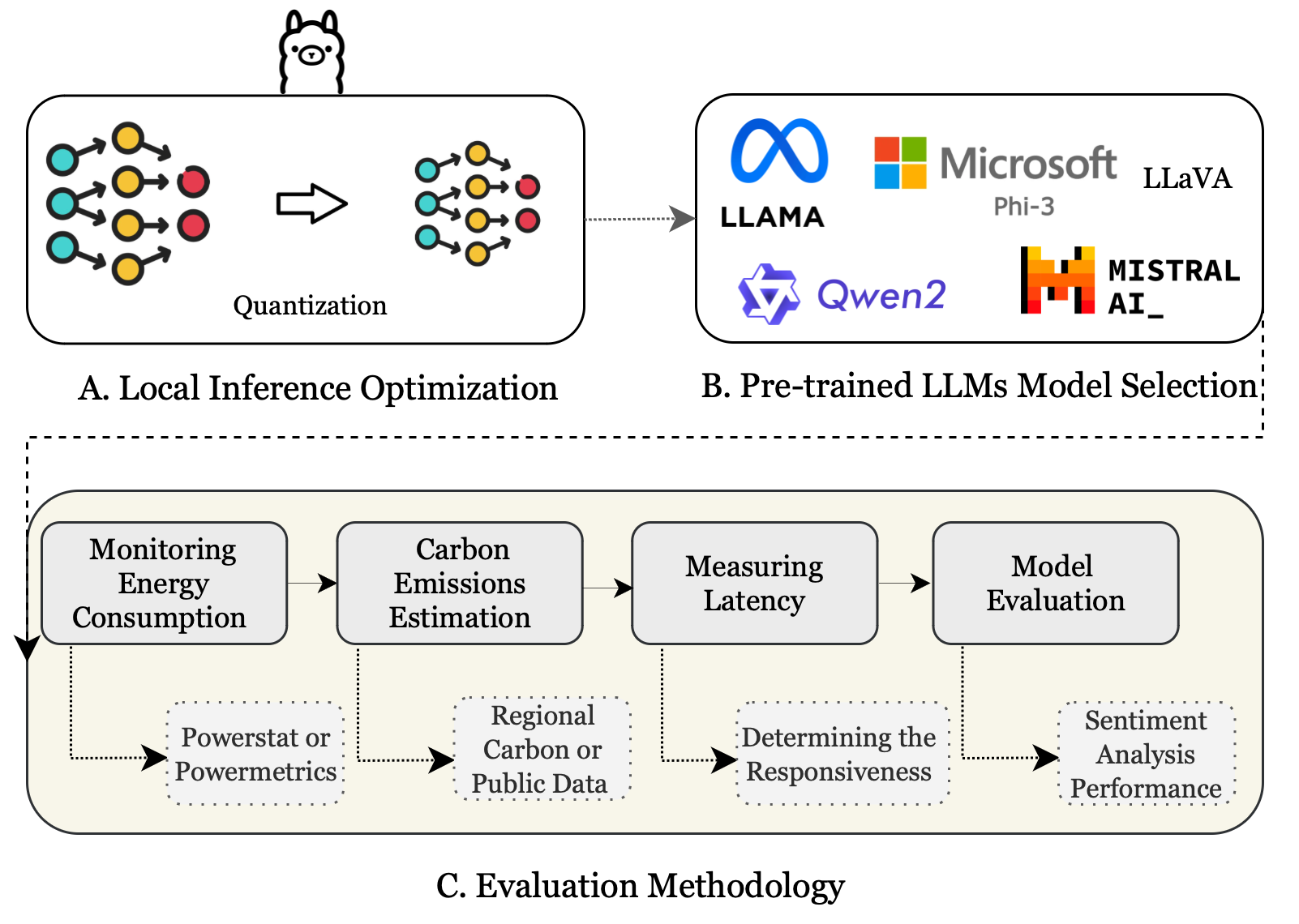}
\caption{Detailed Overview of the Proposed Optimization Framework}
\label{framework_overview}
\end{figure}

The proposed framework (Figure.~\ref{framework_overview}) tackles energy efficiency 
in LLM deployment through three interconnected components: local inference optimization, the selection of energy-efficient LLMs, and a comprehensive evaluation methodology. These components function synergistically to reduce energy consumption without sacrificing predictive accuracy or responsiveness. 

\vspace{-0.5em}

\subsubsection{Local Inference Optimization}
Unlike traditional cloud-based methods that rely on centralized data centers, local inference allows models to run directly on user devices while maintaining data privacy. By minimizing data transmission between clients and remote servers, this method significantly reduces both network overhead and carbon footprint \cite{geens2024energy}. To achieve efficient local inference, the framework employs a \emph{quantization} process \cite{frantar2022gptq}, which lowers the numerical precision of model parameters. Specifically, we define a uniform quantization function \( Q_b(\cdot) \) that maps 32-bit weight tensors to a \(b\)-bit representation:
\begin{equation}
Q_b(w) = \mathrm{round}\!\Bigl(\frac{w - \min(w)}{\Delta}\Bigr)\label{eq},
\end{equation}
where \(\Delta\) is a scaling factor determined by the range (\(\max(w) - \min(w)\)) of the weights. In this work, we use a 4-bit quantization strategy (\(b=4\)), which substantially reduces computational and memory requirements without significantly compromising model performance. We apply quantization through \emph{Ollama} \cite{ollama}, an open-source platform known for its support of edge computing principles and privacy-centric deployments. 

\subsubsection{Selection of Energy-Efficient Pre-trained LLMs}
In addition to local inference optimization, the framework includes a careful selection of pre-trained LLMs that are specifically designed for low computational overhead. These models, including Llama3.2 \cite{llama3.2}, Phi3.2 \cite{phi3.2}, Mistral \cite{mistral7b}, Qwen \cite{qwen2023}, and Llava \cite{llava2023}, stand out for their smaller parameter counts, streamlined architectures, and selective attention mechanisms. Such features align well with edge-oriented design principles, making the models easier to run on devices with limited hardware resources.

\subsubsection{Evaluation Methodology}
The central problem tackled here is a classification task for which we use standard evaluation metrics, including precision, recall, and F1-score. We measure these metrics both \underline{before} and \underline{after} applying our quantization approach to understand any performance trade-offs. Furthermore, we track energy usage and estimate carbon footprints by monitoring power consumption and utilizing emission factor data. Let \( E \) denote the total energy consumed (in kWh), and let \( \alpha \) be the emission factor (kg CO\textsubscript{2} per kWh). We define the carbon footprint \( CF \) as:
\begin{equation}
CF = E \times \alpha.\label{eq},
\end{equation}

\subsection{Expected Outcomes}
The proposed framework is expected to significantly reduce energy consumption and carbon emissions during LLM inference, while maintaining accuracy and responsiveness comparable to standard cloud-based methods. These findings support the goals of \emph{Green AI}, showing that sustainable solutions can deliver high performance without burdening users or compromising model quality.

\section{Experimental Setup}

\subsection{Hardware and Software Setting}
The hardware used includes an 11th Gen Intel(R) Core(TM) i7-1165G7 processor operating at 2.80 GHz (1.69 GHz base frequency), supported by 16.0 GB of installed memory (15.7 GB usable). The system type is a 64-bit operating system with an x64-based processor, running on Windows 11 Pro.

We use Ollama \cite{ollama} for local AI model deployment, which ensures data privacy by processing entirely on-device, ideal for sensitive applications. It supports a variety of pre-trained and fine-tuned models, offering flexibility across use cases. Its lightweight design makes it suitable for both individuals and organizations seeking efficient, secure, and localized AI solutions.


\subsection*{Baselines}
We used the following instruction-tuned models in Table. \ref{tab:baselines} for inference, each configured with specific hyperparameters tailored to their architecture and target tasks. During generation, we restrict the candidate next-token set to control diversity. Top-k keeps only the k most probable tokens at each step, while Top-p (nucleus sampling) keeps the smallest set of tokens whose cumulative probability is at least p. The next token is sampled from this restricted set (with temperature applied, if used). Both parameters are unitless; larger k or p typically increases diversity but can introduce more variance, whereas smaller values approach greedy decoding. The specific settings used here are reported in Table. \ref{tab:baselines}.

\begin{table}[htbp]
\scriptsize
\centering
\caption{Baseline Models and Inference Hyperparameters}
\label{tab:baselines}
\begin{tabular}{|l|c|c|c|c|c|c|}
\hline
\textbf{Model} & \textbf{Batch} & \textbf{Max} & \textbf{Temp.} & \textbf{Top-p} & \textbf{Top-k} & \textbf{Beam} \\ 
\textbf{Name} & \textbf{Size} & \textbf{Tokens} & & & & \textbf{Size} \\ \hline
Llama-3.2-1B & 8 & 512 & 0.7 & 0.9 & 50 & 4 \\ \hline
Phi-3-mini & 8 & 512 & 0.7 & 0.9 & 50 & 4 \\ \hline
Qwen2-7B & 8 & 512 & 0.8 & 0.85 & 40 & 4 \\ \hline
Mistral-7B & 16 & 256 & 0.9 & 0.95 & 30 & 2 \\ \hline
LLaVA-Llama3 & 8 & 512 & 0.7 & 0.9 & 50 & 4 \\ \hline
\end{tabular}
\end{table}

\noindent
The models used are as follows:
\begin{itemize}[leftmargin=*]
\item \textbf{Llama-3.2-1B-Instruct}: An instruction-tuned large language model for general-purpose tasks. 
    \item \textbf{Phi-3-mini-128k-Instruct}: An instruction-tuned multimodal model designed for text and vision integration tasks.
    \item \textbf{Qwen2-7B-Instruct}: A transformer-based language model tuned for general-purpose tasks.
    \item \textbf{Mistral-7B-Instruct-v0.3}: An instruction-tuned model optimized for efficient NLP tasks.
    \item \textbf{LLaVA-Llama3-Instruct}: A fine-tuned version of Llama-3 Instruct with improvements in multiple benchmarks.
\end{itemize}

\subsection{Data}

Our dataset, \textbf{Financial Sentiment Analysis}\cite{malo2014good}, comprises 5,842 entries organized into two columns: \textbf{"text"} and \textbf{"label"}. The \textbf{"text"} column contains the textual data for analysis, while the \textbf{"label"} column indicates the sentiment classification (e.g., positive, negative, or neutral). The dataset is well-structured and contains no missing values, making it highly suitable for sentiment analysis tasks in machine learning studies Figure.~\ref{fig:sentiment_instructions}.

\begin{figure}[htbp]
\small
\begin{tcolorbox}[colback=blue!5!white, colframe=blue!75!black, title=Sentiment Assessment Instructions, sharp corners]
\textbf{Instructions:} Assess the sentiment of the given text by identifying the presence of sentiment indicators such as emotional language, positive or negative expressions, and tone shifts. Mark the sentiment as positive, negative, or neutral and provide reasoning.

\textbf{Text:} {content}

\textbf{Sentiment Indicators Checklist:}
\begin{itemize}
\item \textbf{Emotional Language:} Words that convey strong feelings (e.g., joy, anger, sadness, excitement).
\item \textbf{Positive Expressions:} Words or phrases that promote positive feelings or optimism.
\item \textbf{Negative Expressions:} Words or phrases that express criticism or negativity.
\item \textbf{Tone Shifts:} Noticeable changes in tone that affect how the content is perceived.
\item \textbf{Balanced or Neutral Tone:} The absence of strong emotional language, implying neutrality.
\end{itemize}

\textbf{Response Format:}
\begin{enumerate}[label={}, leftmargin=1cm]
\item \textbf{Positive/Negative/Neutral} [Reasoning]
\item \textbf{Positive/Negative/Neutral} [Reasoning]
\item \textbf{Positive/Negative/Neutral} [Reasoning]
\end{enumerate}
\end{tcolorbox}
\caption{Sentiment Assessment Instructions and Indicators Checklist.}
\label{fig:sentiment_instructions}
\end{figure}
\vspace{-1em}

\section{Results}

\subsection{Accuracy vs Memory Usage}

The results in Table~\ref{tab:optimization_results} show that optimization did not introduce a performance trade-off in this case. Alongside a consistent reduction in carbon emissions across all five models, precision, recall, F1 score, and accuracy all improved after optimization. For example, Llama 3.2 improved from 0.55 to 0.57 in precision and from 0.45 to 0.48 in accuracy, while its CO2 emissions dropped from 0.012 kg to 0.005 kg per inference task. Similarly, Phi 3.2, Qwen, Mistral-small, and Llava-Llama 3 also showed gains across all evaluated performance metrics while achieving lower emissions. These findings suggest that, for this set of models and optimization techniques, quantization and local inference improved both sustainability and predictive performance. Rather than indicating a trade-off, the results demonstrate that optimization can simultaneously enhance model efficiency and effectiveness, although outcomes may still vary depending on the model, task, and deployment context
\begin{table}[ht]
\centering
\caption{Comparison of performance metrics and carbon footprint (CF) for five LLMs before and after optimization. CF is computed as in Equation \eqref{eq} and reported in kg CO$_2$ per inference.
}
\label{tab:optimization_results}
\begin{tabular}{|p{1.8cm}|p{1cm}|p{0.6cm}|p{0.4cm}|p{1cm}|p{1cm}|}
\hline
\toprule
\textbf{Model Name} & \textbf{Precision} & \textbf{Recall} & \textbf{F1} & \textbf{Accuracy} & \textbf{CF} \\ \midrule
\textbf{Before Optimization} & \multicolumn{5}{|c|}{Baseline metrics for comparison} \\ \midrule
Llama 3.2 & 0.55 & 0.45 & 0.44 & 0.45 & 0.012 \\ 
Phi 3.2 & 0.97 & 0.82 & 0.88 & 0.82 & 0.012 \\ 
Qwen & 0.77 & 0.79 & 0.76 & 0.79 & 0.009 \\ 
Mistral-small & 0.70 & 0.67 & 0.65 & 0.67 & 0.020 \\ 
Llava-Llama 3 & 0.58 & 0.50 & 0.48 & 0.50 & 0.014 \\ \midrule
\textbf{After Optimization} & \multicolumn{5}{|p{5cm}|}{Metrics following quantization  and local inference techniques} \\ \midrule
Llama 3.2 & 0.57 & 0.48 & 0.47 & 0.48 & 0.005 \\ 
Phi 3.2 & 1.00 & 0.84 & 0.91 & 0.84 & 0.007 \\ 
Qwen & 0.80 & 0.81 & 0.80 & 0.81 & 0.004 \\ 
Mistral-small & 0.73 & 0.70 & 0.69 & 0.70 & 0.015 \\ 
Llava-Llama 3 & 0.61 & 0.54 & 0.51 & 0.54 & 0.006 \\ 
\bottomrule \hline
\end{tabular}
\end{table}
\vspace{-0.5em}

\subsection{Post-Quantization Performance Evaluation}

The goal of this evaluation is to ensure that, after optimization, predictions remain consistent with ground truth labels and reasoning aligns with the predicted labels. Two subject matter experts assessed predictions based on consistency (alignment with ground truth), clarity (logical and interpretable reasoning), and alignment (agreement between predicted sentiment and reasoning). 
Our results show that most labels and reasoning align well with the model's expectations. We present key examples in Figure.~\ref{Experiments Examples} .

\section{Discussion}

\subsection{Practical impact}
The demonstrated reduction in carbon emissions through optimization techniques such as quantization and local inference holds significant value for industries aiming to enhance sustainability. With models achieving up to 55\% reductions in energy consumption, this work directly aligns with corporate environmental, social, and governance (ESG) goals by lowering operational costs and carbon footprints. These techniques enable the deployment of AI models on edge devices and in resource-constrained environments, expanding their applicability to sectors like IoT, healthcare, and autonomous systems. Industries can leverage this work to build greener AI solutions, reduce reliance on cloud computing, and improve real-time processing capabilities, all while contributing to broader sustainability initiatives.
An additional area for exploration is whether the proposed framework could be extended to support dynamic scheduling based on real-time carbon intensity data. Such integration would allow organizations to further minimize emissions by aligning inference workloads with periods of lower grid carbon intensity, thereby enhancing the environmental benefits of optimized AI deployment.

\subsection{Policy and Regulatory Implications}
Policy frameworks addressing the environmental impact of LLMs have become increasingly critical as their rapid proliferation and computational demands contribute substantially to global energy consumption and carbon emissions. However, the current governance landscape for sustainable LLM development remains fragmented and lacks coherent standards at the international level \cite{henderson2022systematicreportingenergycarbon}. The European Union’s AI Act represents a pioneering attempt to embed sustainability considerations into regulatory design through risk-based classification and environmental impact assessments \cite{EUAIACT}. Additionally, initiatives by the OECD and UNESCO are predominantly normative and non-binding \cite{organisation2022oecd}, \cite{unesco2022recommendation}. It is critical that policymakers integrate sustainability metrics into existing and emerging governance structures through mechanisms such as mandatory carbon disclosure, lifecycle footprint labeling, and energy-efficiency certification. Global initiatives such as the UN’s AI for Good platform and the Standards Development Organizations (SDOs) under the International Telecommunication Union (ITU) can play a pivotal role in fostering agile, multi-stakeholder collaboration to develop sustainability focused technical standards. Embedding such measures within global frameworks would promote transparency, accountability, and innovation in sustainable model optimization techniques such as quantization and local inference. Collectively, these efforts would align LLM development with the United Nations Sustainable Development Goals, particularly SDG 12 (Responsible Consumption and Production) \footnote{https://www.un.org/sustainabledevelopment/sustainable-consumption-production/} and SDG 13 (Climate Action) \footnote{https://www.un.org/sustainabledevelopment/climate-change/}.

\subsection{Limitations}
While optimization techniques such as quantization and local inference deliver substantial reductions in carbon emissions and computational overhead—up to 55\% in our results—their impact on predictive performance is minimal. Metrics such as accuracy, precision, recall, and F1 score remain largely stable, with only slight variations observed across models. This suggests that optimization can be widely adopted without compromising reliability, though applications requiring extremely high precision, such as medical diagnostics or financial modeling, may still warrant additional caution. One consideration is that local inference can place higher demands on hardware, and devices with limited processing power may experience delays in real-time use. Moreover, although quantization reduces model size and improves efficiency, it may introduce numerical instability in certain contexts. These findings highlight that while optimization is a practical and effective pathway to sustainable AI, careful evaluation is still required to ensure robustness across diverse datasets and deployment environments. Future work will further examine system-level effects, dataset characteristics, and model architectures to refine these approaches
Furthermore, the reliance on local inference can lead to slower performance if the underlying hardware is not sufficiently powerful. Devices with limited processing capabilities may experience delays in real-time applications, undermining the efficiency gains achieved through optimization. Additionally, while quantization reduces the size of models, it may introduce numerical instability or rounding errors that could affect the robustness of the predictions, particularly in complex or highly dynamic environments. These limitations highlight the need for careful evaluation of optimization techniques against the specific requirements of a given use case to ensure that the trade-offs are acceptable. Future work will also explore ablation studies to isolate confounding factors such as system-level effects like caching, dataset characteristics, and model architecture.

\begin{figure}[htbp]
\small
\begin{tcolorbox}[colback=white, colframe=green!30, title=Sentiment Evaluation, fonttitle=\bfseries, coltitle=black, sharp corners]
\textbf{Text:} \$AAPL afternoon selloff as usual will be brutal. Get ready to lose a ton of money.\\
\textbf{Label:} Negative \hfill \textbf{Predicted:} Negative\\
\textbf{Reasoning:} ``afternoon selloff as usual will be brutal", ``get ready to lose a ton of money"  strongly negative sentiment.\\
\textbf{Evaluation:} Consistent and correct; matches Negative sentiment with reasoning (e.g., ``brutal" and financial loss).
\end{tcolorbox}

\begin{tcolorbox}[colback=white, colframe=green!30, title=Sentiment Evaluation, fonttitle=\bfseries, coltitle=black, sharp corners]
\textbf{Text:} Dealers said the share was hit by investor disappointment over a refining margin of 9.48 USD per barrel and shipping performance decline.\\
\textbf{Label:} Negative \hfill \textbf{Predicted:} Neutral\\
\textbf{Reasoning:} Negative: ``disappointment" and declining EBIT suggest negativity; Neutral: data tone is factual.\\
\textbf{Evaluation:} Inconsistent prediction; factual tone doesn’t negate ``disappointment" and negative financial performance.
\end{tcolorbox}

\begin{tcolorbox}[colback=white, colframe=green!30, title=Sentiment Evaluation, fonttitle=\bfseries, coltitle=black, sharp corners]
\textbf{Text:} Recent deliveries include refinery technology to Anrak Aluminium and sintering plants to Bhushan Steel.\\
\textbf{Label:} Neutral \hfill \textbf{Predicted:} Neutral\\
\textbf{Reasoning:} Purely factual information, no emotive or evaluative language.\\
\textbf{Evaluation:} Consistent and correct; neutral tone matches factual details.
\end{tcolorbox}

\begin{tcolorbox}[colback=white, colframe=green!30, title=Sentiment Evaluation, fonttitle=\bfseries, coltitle=black, sharp corners]
\textbf{Text:} Cinema Series concludes with a profile of Finnish inventor Olavi Linden, whose work has led to dozens of design awards.\\
\textbf{Label:} Neutral \hfill \textbf{Predicted:} Positive\\
\textbf{Reasoning:} Positive: ``dozens of design awards" implies achievement; Neutral: factual and descriptive tone; ``concludes" is neutral.\\
\textbf{Evaluation:} Inconsistent; factual tone suggests Neutral, despite positive implications of ``awards."
\end{tcolorbox}
\caption{Key Examples of Sentiment Analysis Experiments}
\label{Experiments Examples}
\end{figure}
\noindent

\section{Conclusion}

This paper highlights the critical need for sustainable AI practices, particularly in the deployment of LLMs. By integrating optimization techniques such as quantization and local inference, we successfully demonstrated significant reductions in carbon emissions and energy consumption. The proposed framework provides a practical roadmap for industries and researchers seeking to balance energy efficiency with effectiveness. Future research should explore adaptive optimization strategies to minimize trade-offs and develop new metrics balancing sustainability with predictive performance. Additionally, expanding the framework to address challenges such as numerical instability and task-specific performance degradation will enhance applicability across diverse domains. At a broader level, the findings underscore the necessity of embedding environmental accountability within emerging policy frameworks and international standards governing LLMs. Incorporating environmental impact metrics and energy efficiency standards into governance frameworks is necessary to ensure that sustainability becomes an integral component of regulation rather than an afterthought.


\bibliographystyle{plain}
\bibliography{references}
\end{document}